\UseRawInputEncoding

\pdfoutput=1

\documentclass[11pt]{article}

\usepackage[]{ACL2023}

\usepackage{times}
\usepackage{latexsym}

\usepackage[T1]{fontenc}

\usepackage[utf8]{inputenc}

\usepackage{microtype}

\usepackage{inconsolata}
\usepackage{booktabs}
\usepackage{tabularx}
\usepackage{siunitx}
\usepackage{float}

 \usepackage{multirow}

\newcolumntype{b}{l}
\newcolumntype{m}{>{\hsize=.2\hsize}X}
\newcolumntype{s}{>{\hsize=.25\hsize}X}

\newcommand*{\mynum}[1]{\num[round-mode=places,
                             round-precision=2,
                             propagate-math-font = true,
                             text-series-to-math = true ,
                             group-digits=false]{#1}}

\usepackage{listings}
\usepackage{xcolor}

\definecolor{codegreen}{rgb}{0,0.6,0}
\definecolor{codegray}{rgb}{0.5,0.5,0.5}
\definecolor{codepurple}{rgb}{0.58,0,0.82}
\definecolor{backcolour}{rgb}{0.95,0.95,0.92}

\title{ATLANTIS at SemEval-2025 Task 3: Detecting Hallucinated Text Spans \\ in Question Answering}

\author{{\bf  Catherine Kobus}, {\bf  Francois Lancelot}, \\ {\bf  Marion-Cecile Martin}, {\bf  Nawal Ould Amer} \\
  Airbus AI Research}

\begin{document}
{\makeatletter\acl@finalcopytrue
  \maketitle
}
\begin{abstract}
This paper presents the contributions of the ATLANTIS team to SemEval-2025 Task 3, focusing on detecting hallucinated text spans in question answering systems. Large Language Models (LLMs) have significantly advanced Natural Language Generation (NLG) but remain susceptible to hallucinations, generating incorrect or misleading content. To address this, we explored methods both with and without external context, utilizing few-shot prompting with a LLM, token-level classification or LLM fine-tuned on synthetic data. Notably, our approaches achieved top rankings in Spanish and competitive placements in English and German. This work highlights the importance of integrating relevant context to mitigate hallucinations and demonstrate the potential of fine-tuned models and prompt engineering.
\end{abstract}

\section{Introduction}


LLMs have achieved remarkable proficiency in NLG, enabling significant improvements across various applications, including translation \cite{alves2024toweropenmultilinguallarge}, classification \cite{li2023syntheticdatagenerationlarge}, synthetic data generation \cite{dai2022promptagatorfewshotdenseretrieval}, Retrieval-Augmented Generation (RAG) systems \cite{seo2024retrievalaugmenteddataaugmentationlowresource}. While they have addressed numerous challenges in these domains, they remain prone to hallucination-generating incorrect or misleading content. This issue can undermine system reliability and negatively affect real-world performance, limiting their practical deployment in critical applications.

To tackle this challenge, the SemEval Mu-SHROOM task focuses on detecting hallucinated spans in generated text, a crucial step toward enhancing the trustworthiness of NLG systems. This multilingual task covers 14 languages and requires identifying specific portions of text where hallucinations occur. The task overview paper \cite{vazquez-etal-2025-mu-shroom} provides a comprehensive analysis of the methodologies and findings, offering valuable insights into hallucination detection in NLG systems. The dataset, the evaluation metrics, and models used in the task are also extensively discussed there.

In this paper, we present a set of approaches for the challenge, which fall into two main categories:\\
\noindent  \textbf{Methods without external context} that solely rely on the question and answer as input. We experimented 1) few-shot prompting using LLM and 2) fine-tuning a token-level classifier on our generated synthetic data using MKQA dataset \cite{mkqa}.

\noindent  \textbf{Methods with external context} from Wikipedia, retrieved using our RAG system. This context is then added into our models in three ways: 1) few-shot prompting with LLM,  2) fine-tuning a token-level classifier, and 3) fine-tuning a LLM for hallucinated span detection.

Our models delivered impressive results in several languages. In particular, we ranked first in Spanish, third in English, and fifth in German using few-shot prompting with Gemini Pro, enhanced by contextual information. For French, we achieved the eleventh place with a fine-tuned token classifier model with context.

\section{Systems overview}
In this section, we outline our different approaches. We begin by introducing the retrieval component of our RAG system. Next, we describe our methods based on few-shot prompting with and without retrieval. Finally, we present the approaches that we fine-tuned for the task using synthetic data.

\subsection{Retrieval module}
\label{sec:retrieval_module}

The retrieval module is designed to extract relevant text segments to answer a given question.
We used the Wikipedia dataset\footnote{\url{https://hf.co/datasets/wikimedia/wikipedia}} from November 2023 as source. The text was chunked into segments of 312 tokens each, with an overlap of 100 tokens, resulting in 21 million indexed chunks. These chunks were indexed using both dense representation with BAAI/bge-large-en-v1.5 embedding model\footnote{\url{https://hf.co/BAAI/bge-large-en-v1.5}} and sparse representation with BM25.

The retrieval process comprises three key steps: retrieval, reranking, and clustering.
During the retrieval step, a hybrid search mechanism selects the top 25 chunks. This hybrid search employs both an embedding model and BM25 with distribution-based score fusion \cite{mazzeschi2023}. 
Then, a cross-encoder model\footnote{\url{https://hf.co/BAAI/bge-reranker-large}} reranks these 25 chunks. Based on the computed reranker scores, a k-means clustering algorithm is applied to retain a variable number of the most relevant chunks.

To support multiple languages of the query, we employed a LLM (Mistral-7B-Instruct-v0.2\footnote{\url{https://hf.co/mistralai/Mistral-7B-Instruct-v0.2}}) to translate the query into English. This translation allows the retrieval of relevant Wikipedia context in English from a question in another language.

\subsection{Approaches without additional fine-tuning}

We evaluated two approaches that do not require additional fine-tuning: an overlap-based baseline method and a LLM with a custom prompt.

\subsubsection{Overlap-based method}
The overlap-based method is a heuristic approach that predicts a target token as hallucinated if it does not appear in the context; it is inspired from the overlap-based method detailed in~\cite{DBLP:journals/corr/abs-2011-02593}. Since only the English version of Wikipedia was indexed, this method was tested exclusively for the English language.
\subsubsection{LLM and prompt engineering}

\noindent  \textbf{Prompt Engineering.}
A more flexible approach leverages LLMs, which have demonstrated strong adaptability across various tasks through prompt engineering. To address our challenge, we designed a custom prompt tailored specifically for this task. The LLM was provided with two examples and instructed to generate responses in a structured format—JSON in our case, as illustrated in the prompt \ref{listing:generic_LLM} in the appendix. Our objective was to maximize the capabilities of an LLM by first detecting hallucinations, then implement custom functions to extract the hallucinated spans and identify their positions within the sentence. 

\noindent  \textbf{With retrieval.} After testing fixed prompt strategies, we incorporated textual evidence in the prompt. Providing relevant documents has been shown to reduce hallucination. Moreover, in the challenge setup, it allows for a direct comparison between facts and the answer to be evaluated. The relevant chunks are extracted from Wikipedia English and selected for each question as described in~\ref{sec:retrieval_module}. 
  
To balance between LLM prior knowledge and additional knowledge, rules with different degrees of strictness have been explored, inspired from \cite{wu_2024}. Keeping the flexibility to rely on prior knowledge was important for cases where the retrieval pipeline was unable to find documents with relevant facts or when conflictual information was present in the given chunks. Stricter rules also helped to highlight the minimal hallucinated part in the answer. 
Finally, we tested with including misspellings, such as "Stoveren" instead of "Staveren" for the first example of the English validation set. However, this type of errors was often not labeled in the challenge dataset, therefore we discarded them.

For the first stage of our experiments, we tested on the English dataset only. To adapt to German, French and Spanish, we simply named the language in the prompt and changed one example with question and answer in this other language. Listing \ref{listing:generic_LLM_RAG} shows the prompt used in the multilingual setting. 

\noindent \textbf{Experimental setup.}
The Gemini 1.5 Pro model \cite{gemini152024} was prompted with a fixed seed and a temperature of $0.0$ to foster the replicability of the results. Given the large context size, all the chunks tagged as relevant could be incorporated in the prompt: it represents between 1 and 23 chunks per question, with a median from 4 chunks for French to 6 for Spanish. 

\subsection{Approaches with additional fine-tuning}

The challenge dataset did not provide annotated training data - only small annotated validation dataset. Therefore, we created synthetic data to fine-tune custom models. Using this, we fine-tuned a token-level classifier described in \ref{sec:token_level_classif} and a LLM described in \ref{sec:finetuned_LLM}.

\subsubsection{Data generation process}
\label{sec:data_generation}

LLMs have become increasingly popular for synthetic data generation in various NLP applications \cite{liu2024bestpracticeslessonslearned, seo2024retrievalaugmenteddataaugmentationlowresource}. 
To generate our data, we used MKQA dataset \cite{mkqa}, and excluding long-answer or unanswerable queries.
Given a question and context retrieved using the retrieval module described in~\ref{sec:retrieval_module}, we prompted a LLM (Gemini 1.5 Pro) to generate a short answer and an answer repeating the question to have a format closer to the challenge dataset. Table \ref{table:synthetic_generated_data_1} shows an example. 
Once we get the answer, we prompt the LLM to inject a hallucination, with few-shot learning that provides guidance through examples. The resulting generated dataset contains around 48000 samples (12000 samples per language). Appendix \ref{table:synthetic_generated_data_2} shows two generated samples.

\subsubsection{Token-level classification}
\label{sec:token_level_classif}
\begin{figure*}[htbp]
\centering
\includegraphics[width=1.0\textwidth]{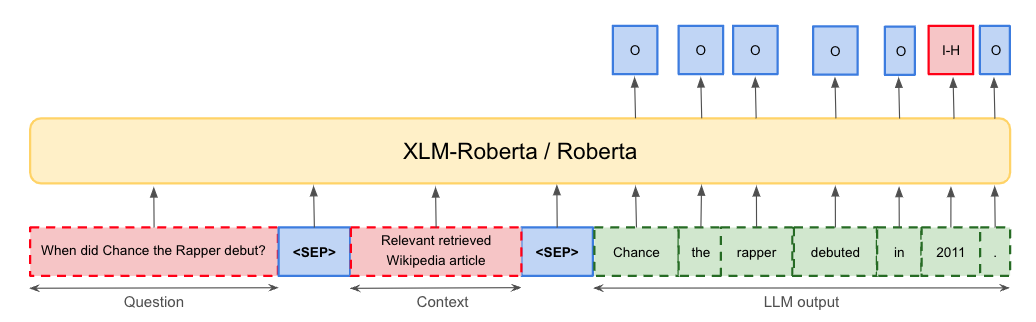}
\caption{Architecture of the token classification approach illustrated with an example}
\label{fig:classification_approach}
\end{figure*}

The span hallucination task can be also casted as a more classical token classification task, where each token in the LLM output is assigned a label, either '\textbf{I-H}' (if the token is part of an hallucination) or '\textbf{O}' (outside an hallucination). This approach takes inspiration from the XLM-R baseline provided by the challenge organizer and from the hallucination detection method for Machine Translation described in~\cite{DBLP:journals/corr/abs-2011-02593}. The architecture of the approach is illustrated in Figure~\ref{fig:classification_approach} with an example taken from the English validation set provided in the challenge. A linear layer is added on top of the pretrained XLM-RoBERTa\footnote{\url{https://huggingface.co/FacebookAI/xlm-roberta-large}} model in order to perform the classification at token level.

We used the synthetic data generated following the procedure detailed in~\ref{sec:data_generation} as training data. Different configurations were tested in the course of the challenge, with or without providing the relevant Wikipedia chunks from ~\ref{sec:retrieval_module}, putting the question either before, after the relevant context or omitting it. For this last configuration, experiments showed that putting the question at the beginning leads to better performances.

Since XLM-RoBERTa has a maximum sequence length of $512$ tokens, we only provide the top-1 retrieved chunk of document as input context to the model. By doing so, the total input length to the model (including, at most, the question, a Wikipedia chunk, and the LLM output) never exceeded the model's maximum sequence length.

During the challenge, different fine-tunings in monolingual and multilingual mode were explored; more details can be found in ~\ref{sec:mono_multilingual}.
    
\noindent \textbf{Experimental setup.} The XLM-RoBERTa large model was fine-tuned for $7$ epochs on $4$ A10G GPUs, with a batch size of $6$ and a learning rate of $2\times10^{-5}$. In the multilingual setting, the training/development data sizes are respectively around $44000$/$4000$ examples, while in the monolingual setting, the training/development data sizes are respectively around $11000$/$1000$ examples.

The checkpoint that gave the best result on the challenge validation set was selected for the test. We also tuned the probability threshold for the '\textbf{I-H}' class. By default, the decision threshold is $0.5$; however, we noticed that the model was globally under-confident, and, by decreasing the threshold, we could increase the results in terms of IoU.

\subsubsection{Fine-tuned LLM}
\label{sec:finetuned_LLM}

We fine-tuned a LLM for hallucination detection, based on the work of \cite{mishra2024finegrained} which introduced FAVA (FAct Vericaton with Augmentation), a model for fine-grained hallucinations detections and editing. We adapted this approach to our question-answering task: we modified the training data to include the question along with the context and answer. Additionally, we simplified the training data by focusing on a single type of hallucinated entity (instead of the six presented in the original paper). We also fine-tuned the LLM with the multilingual synthetic data detailed in~\ref{sec:data_generation} compared to the initial FAVA model which was only fine-tuned for English.

Listing \ref{listing:sample_for_llm_finetuning} shows one sample used for the fine-tuning of the LLM with a French question, the retrieved context in English, and the edited output to correct the hallucination. 

\noindent \textbf{Experimental setup.} We fine-tuned a Llama-3.2-3B-Instruct model for 2 epochs with LoRA, using a batch size of $36$, with $rank=128$ and $\alpha=128$, 4-bit quantization, the Adam optimizer and a learning rate of $2\times10^{-4}$ on $1$ A10G GPU.

\section{Results}

\subsection{Performance of the retrieval system}

Table~\ref{table:retrieval_results} summarizes the retrieval results obtained on the test set. As only the English articles were indexed, we computed the retrieval performance for the other languages by converting the "English" retrieved Wikipedia URL into the target language URL using the MediaWiki API\footnote{\url{https://www.mediawiki.org/wiki/API:Langlinks}} that links equivalent Wikipedia pages in different languages.


\begin{table}[H]\centering
\begin{tabular}{|c|c|c|c|c|} \toprule
    \textbf{language} & \textbf{en} & \textbf{de}& \textbf{es} & \textbf{fr} \\ \midrule
     test & \mynum{0.806}  & \mynum{0.628} & \mynum{0.698} & \mynum{0.624} \\ 
   \bottomrule
\end{tabular}
  \caption{Retrieval scores (MAP@5).}
  \label{table:retrieval_results}
\end{table}
The model performs best in English, with scores (MAP@5) around $0.80$. The lower retrieval performance in other languages (German, Spanish, and French) can be attributed to the additional translation step and the lack of corresponding English-indexed Wikipedia articles for some relevant articles in those languages.

\subsection{Performance of the hallucination detection}
\label{sec:hallucination_results}
Table~\ref{table:hallucination_detection_results} summarizes the results obtained on the test set for four languages: English, German, Spanish and French.

\subsubsection{Performance comparison without retrieval}
In this section, we compare the performance of our strategies in a setting without retrieval, thereby evaluating their standalone capabilities for hallucination detection without external knowledge augmentation.

As reported in Table \ref{table:hallucination_detection_results}, the XLM-RoBERTa large model exhibits moderate performance across languages, achieving its highest test IoU score of $\mynum{0.5025}$ in French, and its lowest in Spanish, $\mynum{0.2652}$. In contrast, Gemini 1.5 Pro demonstrates competitive overall performance, outperforming XLM-RoBERTa in German ($\mynum{0.452}$ vs. $\mynum{0.3848}$) but underperforming in French ($\mynum{0.428}$ vs. $\mynum{0.5025}$).

These findings suggest that, despite the larger scale of general-purpose models like Gemini, smaller models that have been fine-tuned on task-specific data can yield comparable results. Moreover, given that Gemini is pre-trained on general data, our hypothesis is that the fine-tuned XLM-RoBERTa large model would likely exhibit superior performance in domain-specific applications.

\subsubsection{Performance comparison with retrieval}

Here, we focus on the performance of our strategies in a setting with retrieval, thereby evaluating their capabilities for hallucination detection using external knowledge (RAG).
Comparing with the previous section, the scores are better in all languages, without exception. \\
Gemini 1.5 Pro still outperforms the fine-tuned approaches on 3 over 4 languages in this setting. However, the finetuned Llama-3.2-3B reaches the same average IoU of $0.54$ accross all languages, notably given the size difference of these models. Morevover, XLM-RoBERTa, significantly smaller, achieves a score that is relatively close to Llama-3.2-3B in German ($0.53$ vs $0.57$).  \\ These findings suggest that fine-tuning on synthetic data is a promising strategy and leveraging robust retrieval mechanisms with diverse pre-training can yield superior performance in the complex task of hallucinated span extraction.

\begin{table*}[ht]\centering
\begin{tabularx}{\textwidth}{b|mmmm|m} \toprule
     & \textbf{en} & \textbf{de} & \textbf{es} & \textbf{fr} & \textbf{Avg.} \\ \midrule
     \multicolumn{2}{l}{\textbf{Without retrieval}} &   &   &  \\
     Finetuned XLM-RoBERTa large*  &  \mynum{0.4199} &  \mynum{0.3848} & \mynum{0.2652} &  \mynum{0.5025} &  \mynum{0.392} \\
     Gemini 1.5 Pro &  \mynum{0.399} &  \mynum{0.452} & \mynum{0.344} &  \mynum{0.428} &  \mynum{0.405} \\ \bottomrule
     \multicolumn{2}{l}{\textbf{With retrieval}} &   &  &   & \\
     Overlap-based & \mynum{0.359}  & -  & -  & - & -\\

      Finetuned XLM-RoBERTa large\textsuperscript{*}  &  \mynum{0.5146} &  \mynum{0.5276} &  \mynum{0.3693}  & \mynum{0.545}  & \mynum{0.490} \\ 
     Finetuned Llama-3.2-3B\textsuperscript{*} & \mynum{0.552} & \mynum{0.574} & \mynum{0.394}  & \textbf{\mynum{0.628}} &  \mynum{0.535} \\
      Gemini 1.5 Pro &  \textbf{\mynum{0.570}} &  \textbf{\mynum{0.577}}  & \textbf{\mynum{0.531}} &  \mynum{0.495} &  \textbf{\mynum{0.544}}\\ \bottomrule
\end{tabularx}
  \caption{IoU scores on test set. We \textbf{bold} the best performance across submitted systems. Approaches with \textsuperscript{*} were finetuned on a synthetic dataset.}
  \label{table:hallucination_detection_results}
\end{table*}

\subsection{Limitations}
Our approaches have several limitations.

For the retrieval-based method, we indexed only the English version of Wikipedia. If relevant facts reside in other sources, the information retrieval (IR) system cannot provide the necessary context. Additionally, we relied on a LLM to translate queries into English before retrieval. This approach could have been compared with multilingual embedding models and vocabulary-based retrieval to evaluate its effectiveness.

Our prompt-based methods required extensive manual experimentation to design prompts that aligned with the characteristics of this challenge’s dataset. This process was time-consuming and constrained by the limited number of prompts we could test manually. Finding the most effective prompt remains inherently difficult, and a more systematic approach—such as training a model to optimize prompt selection—could have improved our results. 


For the token-level classification model, the limited context window constrained our ability to incorporate all relevant information. Only the first chunk of text was appended to the context, which could be problematic when key details were spread across multiple chunks. A potential solution would be to filter and include only the most relevant sentences to enhance classification accuracy.

Finally, both the token-level classifier and the fine-tuned LLM were trained on synthetic data. Ensuring the accuracy and fidelity of this data is a major challenge. If the synthetic data contains errors, hallucinations, or biases, the trained models may fail to generalize effectively to real-world scenarios, leading to unreliable predictions and reduced robustness \cite{vanbreugel2023syntheticdatarealerrors}. Moreover, the quality of synthetic data depends heavily on the data generation process itself. Addressing these issues would require more rigorous validation techniques or alternative data augmentation strategies to improve the reliability of the training data.

\section{Conclusion}
Different approaches were presented with and without retrieval for the hallucinated span detection task. Overall, the task remains difficult and the performance of the same strategy varies widely depending on the language. This work underlines the importance of adding a relevant context to detect hallucinated spans in the answer. In general, LLM prompting leads to better results and is easily adaptable on other languages but the smaller fine-tuned models show promising results and could thus be preferred, subject to further tuning. Lastly, these approaches would need to be validated on a balanced dataset, containing also a significant part of non-hallucinated answers.

\bibliography{custom}
\bibliographystyle{acl_natbib}

\appendix

\section{Appendix}
\label{sec:appendix}

\subsection{Monolingual/multilingual token-level classification}
\label{sec:mono_multilingual}
This sections contains further details about experiments conducted for the token-level classification strategy, especially with respect to fine-tuning with one or more languages.\\
\textbf{Experimental setup}. At the beginning of the challenge, we focused on English. We finetuned both RoBERTa\footnote{\url{https://huggingface.co/FacebookAI/roberta-large}} and XLM-RoBERTa large models on the English synthetic dataset, with the configurations mentioned in ~\ref{sec:token_level_classif}. The maximum sequence length for RoBERTa is also $512$ tokens, therefore we added the top 1 chunk retrieved when experimenting with retrieval. Then, we decided to extend to other languages of the challenge for which we could create synthetic data - French, Spanish and German. We fine-tuned XLM-RoBERTa in two ways: 

\begin{itemize}
\item Multilingual: on the aggregated synthetic data for all languages
\item Monolingual: for each of the 3 new languages, on the subpart of the synthetic data with the target language
\end{itemize}
For each language, the best checkpoint was selected and the probability threshold was adapted. \\
Table ~\ref{table:roberta_results} shows the IoU scores obtained on the challenge test sets for English, German, Spanish and French. For each configuration, the first score is without adding context, and the second one is with additional context. \\
\textbf{Results}. First of all, the results of ~\ref{sec:hallucination_results} are validated: adding a relevant context always leads to better performances regardless of the finetuning setting.
Monolingual fine-tuning gives higher performance for German (from $0.53$ to $0.55$) and English (from $0.51$ to $0.54$) , whereas better results are reached with multilingual finetuning for French (from $0.46$ to $0.50$) and Spanish (from $0.35$ to $0.37$), with retrieval. In this case, it seems to benefit from the training with data from other languages. \\
The preference to fine-tune specifically on a language or on all varies with respect to the language considered, as well as the performance achieved which is significantly lower for Spanish. Further work could focus on optimizing the fine-tuning to reach a single model that performs well across all these languages. For example, one could use knowledge distillation from the best checkpoints by language into a unique model to obtain multilingual capabilities.
\\


\begin{table*}[ht]\centering
\begin{tabularx}{\textwidth}{ll|llll} \toprule
\textbf{Finetuning setting}    & \textbf{Model}    & \multicolumn{1}{c}{\textbf{en}} & \multicolumn{1}{c}{\textbf{de}} & \multicolumn{1}{c}{\textbf{es}} & \multicolumn{1}{c}{\textbf{fr}} \\ \toprule
\multirow{2}{*}{Monolingual} & Roberta large     & \textbf{\mynum{0.4538}} / \textbf{\mynum{0.5350}}                 & -                               & -                               & -                               \\
                             & XLM-Roberta large & \mynum{0.4525} / \mynum{0.5240}                 & \textbf{\mynum{0.4400}} / \textbf{\mynum{0.5537}}                 & \textbf{\mynum{0.2931}} / \mynum{0.3504}                  & \mynum{0.4617} / \mynum{0.5090}                 \\ \midrule
Multilingual                 & XLM-Roberta large & \mynum{0.4199} / \mynum{0.5146}                 & \mynum{0.3848} / \mynum{0.5276}                 & \mynum{0.2652} / \textbf{\mynum{0.3693}}                 & \textbf{\mynum{0.5025}} / \textbf{\mynum{0.5450}}     \\ \bottomrule
\end{tabularx}
\caption{IoU scores without / with retrieval for the token-level classification strategy on the test set. We \textbf{bold} the best performance across finetuning setups.}
\label{table:roberta_results}
\end{table*}

\subsection{Sample generated data}

\begin{table*}[t]\centering
  \begin{tabularx}{\textwidth}{l|l}
  \toprule
question & when did the first episode of the flash come out\\ \hline
short rag answer & October 7, 2014\\ \hline
rag answer with question & The first episode of The Flash (2014) premiered on October 7, 2014.\\
 \bottomrule
  \end{tabularx}
  \caption{A sample of synthetic generated answer}
  \label{table:synthetic_generated_data_1}
\end{table*}

\begin{table*}[t]\centering
  \begin{tabular}{p{0.30\textwidth}|p{0.60\textwidth}}
  \toprule
question & when did the first episode of the flash come out\\ \hline \hline
short rag answer with hallucination annotations &<entity><mark>October 7, 2014<\/mark><delete>October 7, 2015<\/delete><\/entity>\\ \hline
short rag answer with hallucination &October 7, 2015\\ \hline
mushroom hallucination hard labels & [[0, 15]]\\ \hline \hline
rag answer with question with hallucination annotations & The first episode of The Flash (2014) premiered on <entity><mark>October 7, 2014<\/mark><delete>October 7, 2015<\/delete><\/entity>.\\ \hline
rag answer with question with hallucination &The first episode of The Flash (2014) premiered on October 7, 2015\\ \hline
mushroom hallucination hard labels & [[51, 66]]\\
 \bottomrule
  \end{tabular}
  \center \caption{A sample of synthetic generated hallucination annotation}
  \label{table:synthetic_generated_data_2}
\end{table*}

\subsection{Prompts}

\definecolor{light-gray}{gray}{0.95}
\begin{figure*}[t]
\begin{lstlisting}[caption=Prompt for generic LLM \& multilingual, label={listing:generic_LLM}, language={}, basicstyle=\small, captionpos=b, backgroundcolor = \color{light-gray}]
You are an expert evaluator for language models, tasked with identifying hallucinations or errors in their responses.  A hallucination is defined as:
- Incorrect factual information: Content in the model's response that is unrelated, irrelevant, or factually incorrect with respect to the documents, based on the question.
- Spelling errors: Any word in the model's response that is misspelled or contains typographical mistakes, including incorrect names, places, or other terms.

Your task is to analyze the **Model Output Text** written in {LANGUAGE} and classify hallucinations or errors into two categories:
1. **Factual inaccuracies**
2. **Misspellings**

Ensure the output is in JSON format, following this structure:
'''
{
    "model_input": "<Insert question here>",
    "model_output_text": "<Insert model's response here>",
    "hallucinations": {
        "factual_inaccuracies": [
            "<text_span_1>", "<text_span_2>",  ...
        ],
        "misspellings": [
            "<text_span_3>",
            "<text_span_4>",
            ...
        ]
    }
}
''' 

### Example 1:
**Model Input**: "Quelle est la capitale de la France ?"
**Model Output Text**: "La capitale de la Grance est Berlin."
**Expected JSON Output**:
'''
{
    "model_input": "Quelle est la capitale de la France ?",
    "model_output_text": "La capitale de la Grance est Berlin.",
    "hallucinations": {
        "factual_inaccuracies": [
            "Berlin"
        ],
        "misspellings": [
            "Grance"
        ]
    }
}
### Example 2:
...

### Task:
Now, evaluate the **Model Output Text**. Identify hallucinations or errors, and classify them as either factual inaccuracies or misspellings. The **Relevant Documents** is a list that can contain different documents in English, all independant. If one doesn't seem relevant, don't take it into account to identify hallucinations.

**Model Input**: {QUESTION}
**Model Output Text**: {MODEL_OUTPUT} 

### Remember instruction:
You MUST select only the relevant subparts of the answer (where the error occurs). You MUST split them into the MINIMAL possible parts. You MUST exclude stop words.
\end{lstlisting}
\end{figure*}

\begin{figure*}[t]
\begin{lstlisting}[caption=Prompt for generic LLM - with retrieval \& multilingual, label={listing:generic_LLM_RAG}, language={}, basicstyle=\small, captionpos=b, backgroundcolor = \color{light-gray}]
You are an expert evaluator for language models, tasked with identifying hallucinations or errors in their responses.  A hallucination is defined as:
- Incorrect factual information: Content in the model's response that is unrelated, irrelevant, or factually incorrect with respect to the documents, based on the question.
- Spelling errors: Any word in the model's response that is misspelled or contains typographical mistakes, including incorrect names, places, or other terms.

Your task is to analyze the **Model Output Text** written in {LANGUAGE} and classify hallucinations or errors into two categories:
1. **Factual inaccuracies**
2. **Misspellings**

Ensure the output is in JSON format, following this structure:
'''
{
    "model_input": "<Insert question here>",
    "model_output_text": "<Insert model's response here>",
    "hallucinations": {
        "factual_inaccuracies": [
            "<text_span_1>", "<text_span_2>",  ...
        ],
        "misspellings": [
            "<text_span_3>",
            "<text_span_4>",
            ...
        ]
    }
}
''' 

### Example 1:
**Model Input**: "Quelle est la capitale de la France ?"
**Model Output Text**: "La capitale de la Grance est Berlin."
**Relevant Documents**: ["The following outline is provided as an overview of and topical guide to Paris: Paris - capital and most populous city of France, with an area of and an official estimated population of 2,140,526 residents as of 1 January 2019. Since the 17th century, Paris has been one of Europe's major centres of finance, commerce, fashion, science, and the arts...."]
**Expected JSON Output**:
'''
{
    "model_input": "Quelle est la capitale de la France ?",
    "model_output_text": "La capitale de la Grance est Berlin.",
    "hallucinations": {
        "factual_inaccuracies": [
            "Berlin"
        ],
        "misspellings": [
            "Grance"
        ]
    }
}
### Example 2:
...

### Task:
Now, evaluate the **Model Output Text**. Identify hallucinations or errors, and classify them as either factual inaccuracies or misspellings. The **Relevant Documents** is a list that can contain different documents in English, all independant. If one doesn't seem relevant, don't take it into account to identify hallucinations.

**Model Input**: {QUESTION}
**Model Output Text**: {MODEL_OUTPUT} 
**Relevant Documents**: {CONTEXT}

### Remember instruction:
You MUST select only the relevant subparts of the answer (where the error occurs). You MUST split them into the MINIMAL possible parts. You MUST exclude stop words.
\end{lstlisting}
\end{figure*}

\begin{figure*}[t]
\begin{lstlisting}[caption=Prompt for finetuned LLM, label={table:prompt_finetuned_llm}, language={}, basicstyle=\small, captionpos=b, backgroundcolor = \color{light-gray}]
Given the following question:
{QUESTION}
and the following references:
{CONTEXT}

Identify all the errors in the following passage using the references and the question provided and suggest edits:
{MODEL_OUTPUT}
\end{lstlisting}
\end{figure*}

\begin{figure*}[t]
\begin{lstlisting}[caption=Example of an input/output pair used for the LLM finetuning, label={listing:sample_for_llm_finetuning}, language={}, basicstyle=\small, captionpos=b, backgroundcolor = \color{light-gray}]
USER : Given the following question:
quand le livre de la jungle de Disney est-il sorti
and the following references:
# The Jungle Book (2016 film)
This film required extensive use of computer-generated imagery to portray the other animals and settings. The Jungle Book was released in the United States in Disney Digital 3D, RealD 3D, IMAX 3D, D-Box, and premium large formats, on April 15, 2016. It grossed $966 million worldwide, making it the fifth-highest-grossing film of 2016. Amongst Disney's live-action remakes, this film is the eighth-most-expensive and fifth-highest-grossing readaptation to date. This film won accolades for achievements in visual effects at the 89th Academy Awards, 22nd Critics' Choice Awards, and 7 ...
Identify all the errors in the following passage using the references and the question provided and suggest edits:
Le film est sorti le 18 octobre 1967.

ASSISTANT : Le film est sorti le <entity><mark>15 avril 2016</mark><delete>18 octobre 1967</delete></entity>.
\end{lstlisting}
\end{figure*}

\begin{figure*}[t]
\begin{lstlisting}[caption=Prompt used for translation, label={listing:translation_prompt}, language={}, basicstyle=\small, captionpos=b, backgroundcolor = \color{light-gray}]
Translate the following query into english : {QUERY}. Return only the translated query.
\end{lstlisting}
\end{figure*}

\begin{figure*}[t]
\begin{lstlisting}[caption=Prompt used to generate synthetic hallucination, label={listing:prompt_for_hallucination}, language={}, basicstyle=\small, captionpos=b, backgroundcolor = \color{light-gray}]
You must follow those examples to generate a new answer with annotations.

Example 1 :
- Question :
What did Petra van Staveren win a gold medal for?
- Answer :
Petra van Stoveren won a gold medal in the 1984 Summer Olympics in Los Angeles, USA.
- Answer with annotations :
Petra van Stoveren won a <entity><delete>gold</delete><mark>silver</mark></entity> medal in the <entity><delete>1984</delete><mark>2008</mark></entity> Summer Olympics in <entity><delete>Los Angeles, USA</delete><mark>Beijing, China</mark></entity>.

Example 2 :
- Question :
Which network released the TV series of the The Punisher?
- Answer :
The Punisher network that released this TV show is Netflix.
- Answer with annotations :
The <entity><delete>Punisher</delete><mark>Puncher</mark></entity> network that released this TV show is Netflix.
...
Example :
- Question :
{QUESTION}
- Answer :
{ANSWER}
- Answer with annotations :
\end{lstlisting}
\end{figure*}

\end{document}